\documentclass[conference]{IEEEtran}
\ifCLASSINFOpdf
\usepackage[pdftex]{graphicx}
\DeclareGraphicsExtensions{.pdf,.jpeg,.png}
\else
\usepackage[dvips]{graphicx}
\DeclareGraphicsExtensions{.eps}
\fi
\usepackage[cmex10]{amsmath}
\usepackage{amssymb}
\usepackage{amsthm }
\usepackage{nicefrac}
\usepackage{hyperref}
\hypersetup{colorlinks=true} 
\usepackage{xcolor}
\usepackage{tabularx}

\theoremstyle{remark} 
 
\newcommand{\avec}{{\bf{a}}}

\newcommand{\cvec}{{\bf{c}}}

\newcommand{\svec}{{\bf{s}}}

\newcommand{\xvec}{{\bf{x}}}
\newcommand{\yvec}{{\bf{y}}}

\newcommand{\zerovec}{{\bf{0}}}
\newcommand{\onevec}{{\bf{1}}}

\newcommand{\Real}{\mathbb{R}}

\newcommand{\etab}{{\mbox{\boldmath $\eta$}}}

\newcommand{\Cb}{\mathbf{C}}
\newcommand{\Db}{\mathbf{D}}
\newcommand{\Eb}{\mathbf{E}}
\newcommand{\Hb}{\mathbf{H}}
\newcommand{\Ib}{\mathbf{I}}
\newcommand{\Kb}{\mathbf{K}}

\newcommand{\Ob}{\mathbf{O}}

\newcommand{\Rb}{\mathbf{R}}
\newcommand{\Sb}{\mathbf{S}}

\newcommand{\Ub}{\mathbf{U}}
\newcommand{\Wb}{\mathbf{W}}
\newcommand{\Xb}{\mathbf{X}}
\newcommand{\Yb}{\mathbf{Y}}

\newcommand{\Gammab}{\mathbf{\Gamma}}

\newcommand{\norm}{\vert \vert}
\newcommand{\lag}{{\cal L}}
\newcommand{\lzero}{$\ell_{0}$}
\newcommand{\lone}{$\ell_{1}$}
\newcommand{\ltwo}{$\ell_{2}$}
\newcommand{\CSIM}{{\mathrm{CSIM}}}
\newcommand{\dotp}{{\odot }}

\usepackage{algorithm}
\usepackage{algorithmic}
\usepackage{multirow}
\usepackage[bottom]{footmisc}
\ifCLASSOPTIONcompsoc
  \usepackage[caption=false,font=normalsize,labelfont=sf,textfont=sf]{subfig}
\else
  \usepackage[caption=false,font=footnotesize]{subfig}
\fi
\usepackage{fixltx2e}
\usepackage{afterpage}
\usepackage{float}
\usepackage{stfloats}
\begin{document}

%
\title{Recovery of Missing Samples Using Sparse Approximation via a Convex Similarity Measure}

\author{\IEEEauthorblockN{Amirhossein Javaheri}
\IEEEauthorblockA{Department of Electrical Engineering\\
Sharif University of Technology\\
Tehran, Iran\\
Email: javaheri\_amirhossein@ee.sharif.edu}
\and
\IEEEauthorblockN{Hadi Zayyani}
\IEEEauthorblockA{Departement of Electrical and\\
Computer Engineering\\
Qom University of Technology\\
Qom, Iran\\
Email: zayyani@qut.ac.ir}
\and
\IEEEauthorblockN{Farokh Marvasti}
\IEEEauthorblockA{Department of Electrical Engineering\\
Sharif University of Technology\\
Tehran, Iran\\
Email: marvasti@sharif.edu}}



%


\maketitle
\thispagestyle{plain}
\pagestyle{plain}
\begin{abstract}
In this paper, we study the missing sample recovery problem using methods based on sparse approximation. In this regard, we investigate the algorithms used for solving the inverse problem associated with the restoration of missed samples of image signal. This problem is also known as inpainting in the context of image processing and for this purpose, we suggest an iterative sparse recovery algorithm based on constrained $l_1$-norm minimization with a new fidelity metric. The proposed metric called Convex SIMilarity (CSIM) index, is a simplified version of the Structural SIMilarity (SSIM) index, which is convex and error-sensitive. The optimization problem incorporating this criterion, is then solved via Alternating Direction Method of Multipliers (ADMM). Simulation results show the efficiency of the proposed method for missing sample recovery of 1D patch vectors and inpainting of 2D image signals.
\end{abstract}


%
\IEEEpeerreviewmaketitle

\section{Introduction}
\IEEEPARstart{T}{he} algorithms for missing sample recovery, have many applications in the field of signal and image processing \cite{Craig}, where in the latter it is specifically known as image inpainting \cite{Guil14}. 
Among various methods for missing sample recovery and inpainting, including diffusion-based \cite{Chan01} and exemplar-based \cite{Casellas11} methods, some exploit the sparsity of the signals in the transform domain \cite{Guler06,Elad05,Fadili07,Li14}. In this paper, we study this class of algorithms where in, it is assumed that the signal is sparse within a discrete transform domain. The sparsity of the signal, enables us to reconstruct it from random measurements even below the rate of Nyquist. This is known as Compressed Sensing (CS) \cite{Dono06} which has applications in different areas in signal processing \cite{Marv12}. The problem of reconstruction of the sparse signal from random samples, is also a problem of inverse modelling. Many algorithms are introduced for this purpose within different applications in audio and image processing.

Methods based on spare approximation use an index as the measure of sparsity which is stated in terms of the $p$-norm of the sparse signal. The most common vector norms used to promote sparsity are \lzero~and \lone~and the algorithms for sparse recovery can be generally divided into two groups; The algorithms based on \lzero-minimization and those based on \lone-minimization or the basis-pursuit method \cite{MallZ93}. For detailed survey on sparse recovery algorithms, see \cite{Zhang15}. 

In this paper, we propose an alternative \lone-minimization method for sparse recovery of image signals. The proposed method has application in image inpainting and restoration. We introduce a fidelity criterion called Convex SIMilarity (CSIM) Index, which has desirable features including convexity and error-sensitivity. We study the missing sample recovery for 1D and 2D signals using the proposed index as fidelity criterion in our optimization problem. The 1D recovery algorithm is applied for reconstruction of the vectorized small patches, whereas its 2D variant can be directly used to recover missing samples of the entire image signal.

\section{Image Quality Assessment}
\label{sec:QualityAssessment}
There are different criteria for Image Quality Assessment (IQA). The most popular metric is \ltwo~norm or MSE. But, there are cases in which the MSE criterion fails to to accurately recover the original signal. One reason is that this metric is insensitive to the distribution of the error signal. Thus, there are a class of alternative perceptual criteria introduced for error-sensitive visual quality assessment. The most popular measure from this class is SSIM, defined as \cite{SSIM04}:
\begin{equation}
\label{eq_SSIM}
\mathrm{SSIM}(\xvec,\yvec) =\Big( \dfrac{2\mu_x \mu_y+C_1}{\mu_x^2+\mu_y^2+C_1} \Big) \Big(
\dfrac{2\sigma_{x,y}+C_2}{\sigma_x^2+\sigma_y^2+C_2} \Big)
\end{equation}
where $C_1, C_2>0$ are constant. This function whose mathematical properties are discussed in \cite{SSIM12}, is non-convex (within its entire definition domain), implying that its global optimization is complex. Hence, in this paper, we use a simplified criterion derived from the numerator and the denominator terms appearing in \eqref{eq_SSIM}.
The proposed index named CSIM is defined as follows:
\begin{equation}
\label{eq_1}
\mathrm{CSIM}(\xvec, \yvec) = K_0\Big((\mu_x^2 + \mu_y^2 - 2\mu_x \mu_y)+\rho (\sigma_x^2+\sigma_y^2-2\sigma_{x,y})\Big)
\end{equation}
where $\rho$ and $K_0$ are positive constants. The first parameter controls sensitivity with respect to random disturbances versus uniform change, i.e., unlike MSE, the new criterion has error-sensitive variation. The parameter $K_0$ is just used for scaling.
The proposed index also has feasible mathematical features, including convexity (uni-modality) and positive-definiteness. It can also be shown that if we use statistical estimates for mean and variance/covariance of signals $\xvec, \yvec \in \Real ^n$, the function defined in \eqref{eq_1} is algebraically equivalent to:
\begin{equation}
\label{eq_8}
 \CSIM(\xvec, \yvec) = (\xvec-\yvec)^T\Wb (\xvec-\yvec)
\end{equation}
where $\Wb_{n\times n} =w_1 \Ib_n+ w_2\onevec_n \onevec_n^T$ ($\onevec_n = (1,\ldots, 1)_{1 \times n}^T$) and $w_1$ and $w_2$ are obtained as:
\begin{equation}
\label{eq_theta}
w_1=\frac{K_0\rho}{n-1}, \quad w_2=K_0 \left( \frac{1}{n^2}-\frac{\rho}{n(n-1)}\right)
\end{equation}

\section{The problem formulation}
\label{sec:Problem}
Suppose $\xvec \in \mathbb{R}^N$ is the vectorized image signal and $\Hb\in \Real^{m\times N}$ is the random sampling matrix which determines the pattern of the available (missed) samples. The observed signal with missing samples, is also denoted by $\yvec= \Hb \xvec \in \Real^m$. If we assume that $\xvec$ has approximately a sparse representation via a dictionary basis specified by $\Db$, the regular optimization problem for sparse recovery is as follows:
\begin{equation}
\label{eq_19}
\min_{\svec} \norm \svec \norm_1 \quad \textit{s.t.} \; 
\norm \Hb\Db \svec-\yvec \norm_2^2 \leq \epsilon_n
\end{equation}
where $\svec$ denotes the sparse vector of representation coefficients and $\epsilon_n$ denotes the upper bound for the energy of the additive noise (in case the observed signal is noisy). 
In addition to \ltwo~norm fidelity criterion, there are also methods based on perceptual image quality metrics for recovery of the missing samples. In \cite{Oga13} a method for image completion is proposed, where for each corrupted patch, the optimization problem below is proposed: 

\begin{equation}
\label{eq_18_9}
\max_{\xvec,\svec}
\mathrm{SSIM}(\xvec, \Db \svec) \quad \textit{s.t.} \; \left\{\begin{IEEEeqnarraybox}[\relax][c]{l}
 \Hb \xvec=\yvec \\
\norm \svec\norm_0 \leq T%
\end{IEEEeqnarraybox}\right.
\end{equation} 
This problem is iteratively solved using an approach based on Matching Pursuit \cite{MallZ93} via quasi-linear search methods proposed in \cite{Chann08}.

\section{The proposed algorithm}
\subsection{$\mathrm{1D}$ sparse recovery}
\label{sec:1DSparseRecovery}
As discussed earlier, most of the algorithms use \ltwo~norm as fidelity criterion for image reconstruction. Here, we propose to use CSIM instead of MSE to solve the missing sample recovery problem \eqref{eq_19}. Hence, by introducing the auxiliary variable $\xvec$, we have the following optimization problem:
\begin{equation}
\label{eq_19_2}
\min_{\xvec,\svec} \norm \svec \norm_1 \quad \textit{s.t.} \; \left\{\begin{IEEEeqnarraybox}[\relax][c]{l}
 \CSIM(\Hb\xvec,\yvec ) \leq \epsilon_n \\
\xvec=\Db \svec%
\end{IEEEeqnarraybox}\right.
\end{equation}
Now, using Lagrange multipliers theorem, satisfying Karush-Kuhn-Tucker (KKT) conditions \cite{LagBertsekas}, the optimization problem above is equivalent to:
\begin{equation}
\label{eq_20}
\min_{\xvec,\svec}
  \CSIM(\Hb\xvec,\yvec)+\alpha \norm \svec \norm_1 \quad \textit{s.t.}\; \xvec=\Db \svec
\end{equation}
The convexity property of $\CSIM$ gives the guarantee to use the Alternating Direction Method of Multipliers (ADMM) \cite{BoydADMM} to solve \eqref{eq_20}. Hence, the augmented Lagrangian cost function is:
\begin{align}
\label{eq_22}
\min_{\xvec,\svec}  \, \lag (\xvec,\svec)= & \, \CSIM(\Hb\xvec,\yvec) +\alpha \norm \svec \norm_1  +\etab^T (\xvec-\Db \svec) \nonumber \\  
& + \frac{\sigma}{2} \norm\xvec-\Db \svec \norm_2^2 
\end{align}
The ADMM alternatively minimizes \eqref{eq_22} with respect to each variable. Hence, at each iteration of the ADMM, the problem \eqref{eq_22} is separated into sub-problems as follows:
\subsection{$\xvec$ sub-problem:}
\label{subsec:x}
At $t$-th iteration, the augmented Lagrangian function associating with $\xvec$ is:
\begin{equation}
\label{eq_23}
\xvec^{(t+1)}=\mathop{\mathrm{argmin}}_{\xvec}
\lag (\xvec,\svec^{(t)}) = \frac{1}{2}\xvec^T \Kb\xvec+ {\cvec ^{(t)}}^T \xvec
\end{equation}
where 
\begin{align}
\label{eq_23_2}
\Kb &=2 \Hb^T \Wb \Hb +\sigma \Ib \nonumber \\
{\cvec ^{(t)}}&=-2\Hb^T \Wb \yvec -\sigma \Db \svec^{(t)}+\etab^{(t)}
\end{align} 
This quadratic problem has a closed-form solution:
\begin{equation}
\label{eq_24}
\xvec^{(t+1)} =-\Kb^{-1} \cvec^{(t)} =-\frac{1}{\sigma} \Big[ \Hb^T({2}/{\sigma}\Wb)\Hb + \Ib\Big]^{-1} \cvec^{(t)}
\end{equation}
Now, using the Sherman-Morrison-Woodbury lemma \cite{Woodbury} to calculate the inverse matrix, we have:
\begin{equation}
\label{eq_24_2}
\Big[\Hb^T({2}/{\sigma}\Wb)\Hb + \Ib\Big]^{-1}\!\!\!\!  = \Big[ \Ib - \Hb^T(\gamma_1 \Ib + \gamma_2 \onevec \onevec^T) \Hb\Big]
\end{equation}
where 
\begin{equation}
\gamma_1 = \frac{1}{\beta_1},\quad \gamma_2 = -\frac{\beta_2}{\beta_1(\beta_1 +n\beta_2)}
\end{equation}
and $\beta_1 = \frac{\sigma}{2 w_1}+1, \quad \beta_2=- \frac{\sigma }{2w_1}\frac{w_2}{w_1+n w_2}$. The parameters $w_1, \,w_2$ are also given in \eqref{eq_theta} and equation \eqref{eq_24_2} yields by applying the inverse matrix lemma consecutively to obtain $(\beta_1 \Ib +\beta_2  \onevec \onevec^T)^{-1}$. We have also used $\Hb \Hb^T = \Ib$. Thus the update formula for $\xvec$ is:
\begin{equation}
\label{eq_25}
\xvec^{(t+1)} =-\frac{1}{\sigma} \left(\Ib-\gamma_1 \Hb^T \Hb-\gamma_2 \onevec_H  \onevec_H^T \right) \cvec^{(t)}, \quad \onevec_H = \Hb^T \onevec
\end{equation}
\subsection{$\svec$ sub-problem:}
\label{subsec:s}
The optimization sub-problem associating with $\svec$ is:
\begin{equation}
\label{eq_26}
\svec^{(t+1)}=\mathop{\mathrm{argmin}}_{\svec} \lag (\xvec^{(t+1)},\svec) = \mathop{\mathrm{argmin}}_{\svec}\lag(\svec)
\end{equation}
Using the Majorization Minimization (MM) technique \cite{LangeMM}, assuming $\norm \Db \norm_2^2 = \lambda$,  Similar to \cite{Daub04} we define the surrogate function $\lag^S(\svec,\svec_0)=\lag(\svec)+\frac{\lambda}{2} \norm \svec_0-\svec \norm_2^2 -\frac{1}{2} \norm\Db\svec_0-\Db \svec \norm_2^2$.

After simplifications, we have:
\begin{equation}
\label{eq_28}
\min_{\svec}\lag^S(\svec,\svec_0) =\min_{\svec} \,\frac{\lambda}{2} \norm \svec - \avec(\svec_0) \norm_2^2 + \frac{\alpha}{\sigma} \norm \svec \norm_1
\end{equation}
where $\avec(\svec_0)=\svec_0+\frac{1}{\lambda } \Db^T \big(\xvec^{(t+1)}-\Db \svec_0+\frac{1}{\sigma}\etab^{(t)} \big)$. The solution to \eqref{eq_28} is obtained using the soft-thresholding operator ${\cal S}$ \cite{Daub04}.
Setting $\svec_0=\svec^{(t)}$ we get:
\begin{align}
\label{eq_29}
\svec^{(t+1)} ={\cal S}_{\!{\frac{\alpha}{\lambda\sigma}} } \Big( \svec^{(t)}+\frac{1}{\lambda} \Db^T \big(\xvec^{(t+1)}+\frac{1}{\sigma}\etab^{(t)}-\Db \svec^{(t)}\big) \Big)
\end{align}
Now since $\svec^{(t+1)}$ is the minimizer of $\lag^S(\svec,\svec^{(t)})$ and $\lag^S(\svec,\svec^{(t)}) \geq \lag(\svec), \, \forall \svec \neq  \svec^{(t)}$, minimizing \eqref{eq_28} will reduce the initial cost function $\lag(\svec)$.
We also use the exponential thresholding method proposed in \cite{Marv003} to decrease the regularizing parameter $\alpha$ accordingly.

\subsection{$\mathrm{2D}$ sparse recovery}
\label{sec:21SparseRecovery}
The proposed algorithm for patch vector reconstruction is given in Algorithm \ref{Algorithm_1}. This method can also be directly applied for holistic 2D image signal recovery from random samples. If we denote the image signal by the matrix $\Xb\in \Real^{n_1\times n_2}$, and the sparsifying 2D transform by $\mathrm{DT2D}$, we have:
\begin{algorithm}[!t]
\caption{Proposed algorithm for 1D missing sample \newline recovery (To solve problem \eqref{eq_22})}
\textbf{Input} $\yvec = \Hb \xvec_0$ and $\Hb$ ($\xvec_0$ is the original signal)\newline
\textbf{Set}   $\sigma>0,\, K_0>0,\, \rho\geq 1,\, \mu<1,\, \zeta<1,\, \alpha_{\min}\ll1$,\newline
\textbf{Initialize} $\alpha=\zeta \norm \Db^T \yvec \norm _{\infty}$, $\etab^{(0)}=\zerovec$, $\svec^{(0)}=\zerovec$, $t=0$.
\begin{algorithmic}[1]
\REPEAT
\STATE Obtain $\cvec^{(t)}$ and update $\xvec^{(t+1)}$ using \eqref{eq_23_2} and \eqref{eq_25}
\STATE
Update $\svec^{(t+1)}$ using \eqref{eq_29}
\STATE Update $\etab^{(t+1)}=\etab^{(t)}+\sigma (\xvec^{(t+1)}-\Db\svec^{(t+1)})$ 
\STATE Update $\alpha= \max(\alpha\times\mu, \alpha_{\min})$.
\STATE $t \leftarrow t+1$
\UNTIL {A stopping criterion is reached}
\end{algorithmic}
\textbf{Output}  $\hat{\xvec}=(\Ib-\Hb^T\Hb)\xvec^{(t_{\text{end}})}+\Hb^T\yvec $
\label{Algorithm_1}
\end{algorithm}

\begin{equation}
\label{eq_DT2D}
\Sb = \mathrm{DT2D}(\Xb), \quad \norm \Sb\norm_{1,1}=\sum_{i} \sum_j \vert s_{i,j} \vert \leq T
\end{equation}
where $\Sb$ denotes the sparse matrix of the transform coefficients. If the inverse transform exists, we have $\Xb = \mathrm{IDT2D}(\Sb) \equiv \Db \svec$, and thus, $\mathrm{IDT2D}$ may be considered a 2D basis. Taking advantage of this, we can harness the sparsity of fast 2D transforms like DCT2D and Curvelet \cite{Curv}, which exploit 2D dependencies between pixels in an image more efficiently compared to 1D transforms. First of all, for a 2D signal, the CSIM criterion is reformulated as:

\begin{algorithm}[!b]
\caption{Proposed algorithm for 2D image inpainting\newline using 2D sparsifying transform}
\textbf{Input} $\Yb=\Hb\dotp \Xb_0$ and $\Hb$ ($\Xb_0$ is the original image) \newline
\textbf{Set}   $\sigma>0,\, K_0>0,\, \rho\geq 1,\,\mu<1,\, \zeta<1,\,\alpha_{\min}\ll1$,\newline
\textbf{Initialize} $\alpha=\zeta \max(\max(\vert \mathrm{DT2D}(\Yb) \vert))$, $\Gammab^{(0)}=\Ob$,\newline $\Sb^{(0)}=\Ob$, $\Ub^{(t)}=\Ob$, $t=0$.

\begin{algorithmic}[1]
\REPEAT
\STATE Obtain $\Cb^{(t)}=\! -2\Wb (\Yb\dotp \Hb)- \sigma\Ub^{(t)} +\Gammab^{(t)}$ 
\STATE Update \\
$\Xb^{(t+1)}\!= \!-\frac{1}{\sigma}\Big[ \!\Cb^{(t)}-\gamma_1 \Cb^{(t)} \dotp \Hb -\gamma_2 \Wb(\Cb^{(t)} \dotp \Hb) \dotp \Hb\Big]$

\STATE
Obtain $\Rb^{(t)}= \mathrm{Intp}\Big(\Xb^{(t+1)}+\frac{1}{\sigma}\Gammab^{(t)} -\Ub^{(t)})\Big)$ 
\STATE
Update $\Sb^{(t+1)}={\cal S}_{\frac{\alpha}{\lambda \sigma}} \Big(\mathrm{DT2D}\big(\Ub^{(t)}+ \frac{1}{\lambda} \Rb^{(t)}\big) \Big)$ 
\STATE Update $\Ub^{(t+1)}=\mathrm{IDT2D}(\Sb^{(t+1)})$
\STATE Update $\Gammab^{(t+1)}=\Gammab^{(t)}+\sigma (\Xb^{(t+1)}-\Ub^{(t+1)})$ 
\STATE Set $\alpha= \max(\alpha\times\mu, \alpha_{\min})$.
\STATE $t \leftarrow t+1$
\UNTIL {A stopping criterion is reached}
\end{algorithmic}
\textbf{Output}  $\hat{\Xb}=(\Ib-\Hb)\dotp\Xb^{(t_{\text{end}})}+\Hb\dotp\Yb $
\label{Algorithm_2}
\end{algorithm}

\begin{equation}
\CSIM (\Xb,\Yb) = \mathrm{trace}\Big( w_1 \Eb^T\Eb +w_2 \onevec_{n_1} ^T \Eb \onevec_{n_2} \onevec_{n_2}^T \Eb^T\onevec_{n_1} \Big)
\end{equation}
where $\Eb = \Xb-\Yb$. Furthermore, the product $\Wb \xvec$ used in the first step of the proposed algorithm ($\xvec$ sub-problem), in matrix form, is equivalent to the function below:  
\begin{equation}
\Wb (\Xb) =w_1 \Xb + w_2  \onevec_{n_1}  \onevec_{n_1} ^T \Xb \onevec_{n_2} \onevec_{n_2}^T
\end{equation}
We can also model the sampling process with component-wise (Hadamard) dot product by a 2D binary sampling mask $\Hb\in \Real ^{n_1 \times n_2}$, i.e., $\Hb \Xb = \Hb \dotp \Xb$.
In addition to these 1D-2D conversions (modifications), we also exploit an idea to improve the performance of the algorithm for inpainting. 
As proposed in \cite{Azgh13}, in each $\Sb$ upgrading step, we apply a linear interpolation method to provide a more enhanced estimate of the residual image. Hence, the final method is given in Algorithm \ref{Algorithm_2}  
where $\mathrm{Intp}$ denotes the linear interpolation function, performed by moving average filtering the input signal\footnote{MATLAB command \texttt{im2filter}}.

\begin{figure*}[!t]
\vspace{-1cm}
\hspace{-0.6cm}
\subfloat[PSNR (DCT 128)]{\includegraphics[width=2.0in]{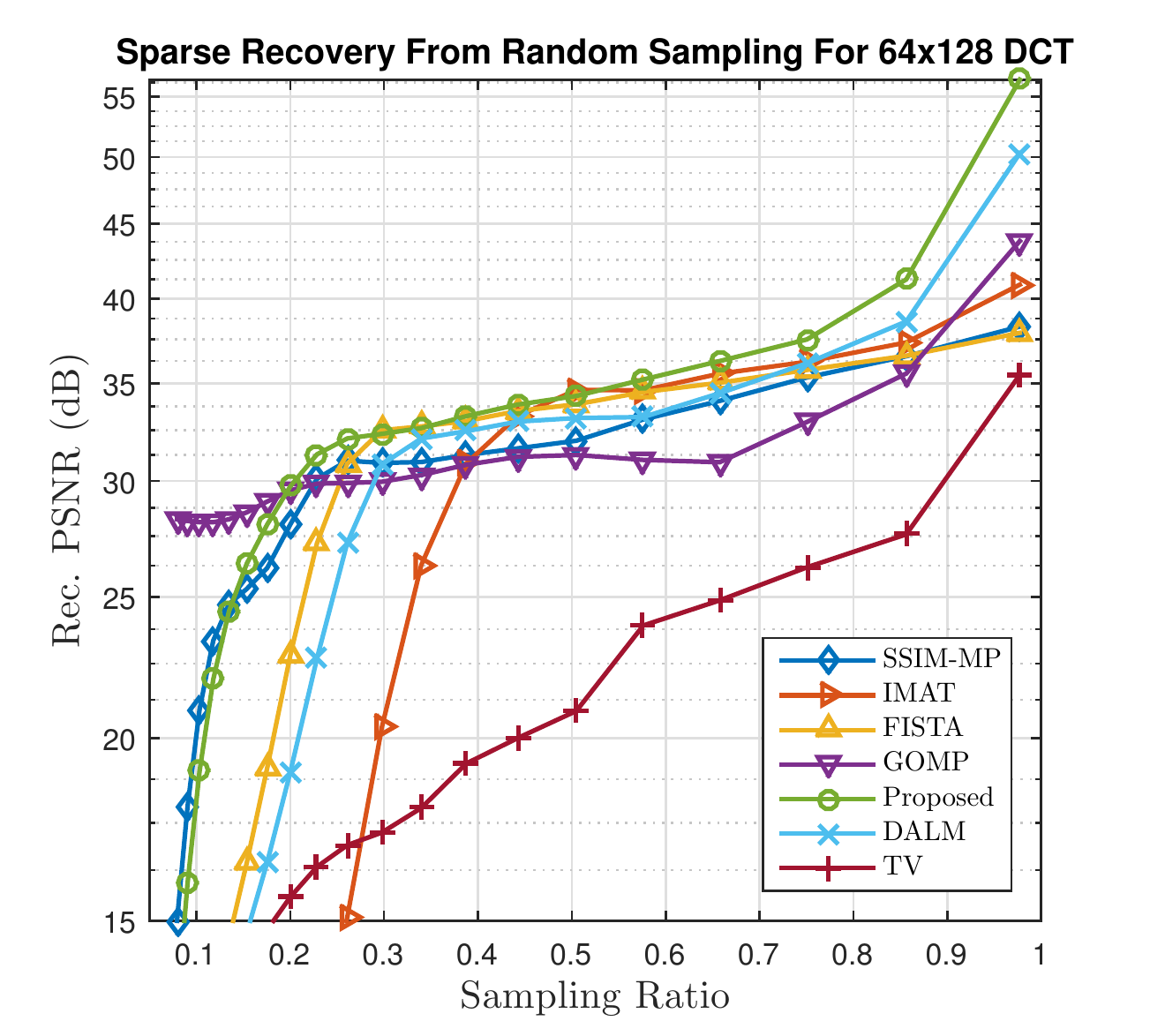}%
\label{fig_1:a}}  \hspace{-0.4cm}
\subfloat[SSIM (DCT 128)]{\includegraphics[width=2.0in]{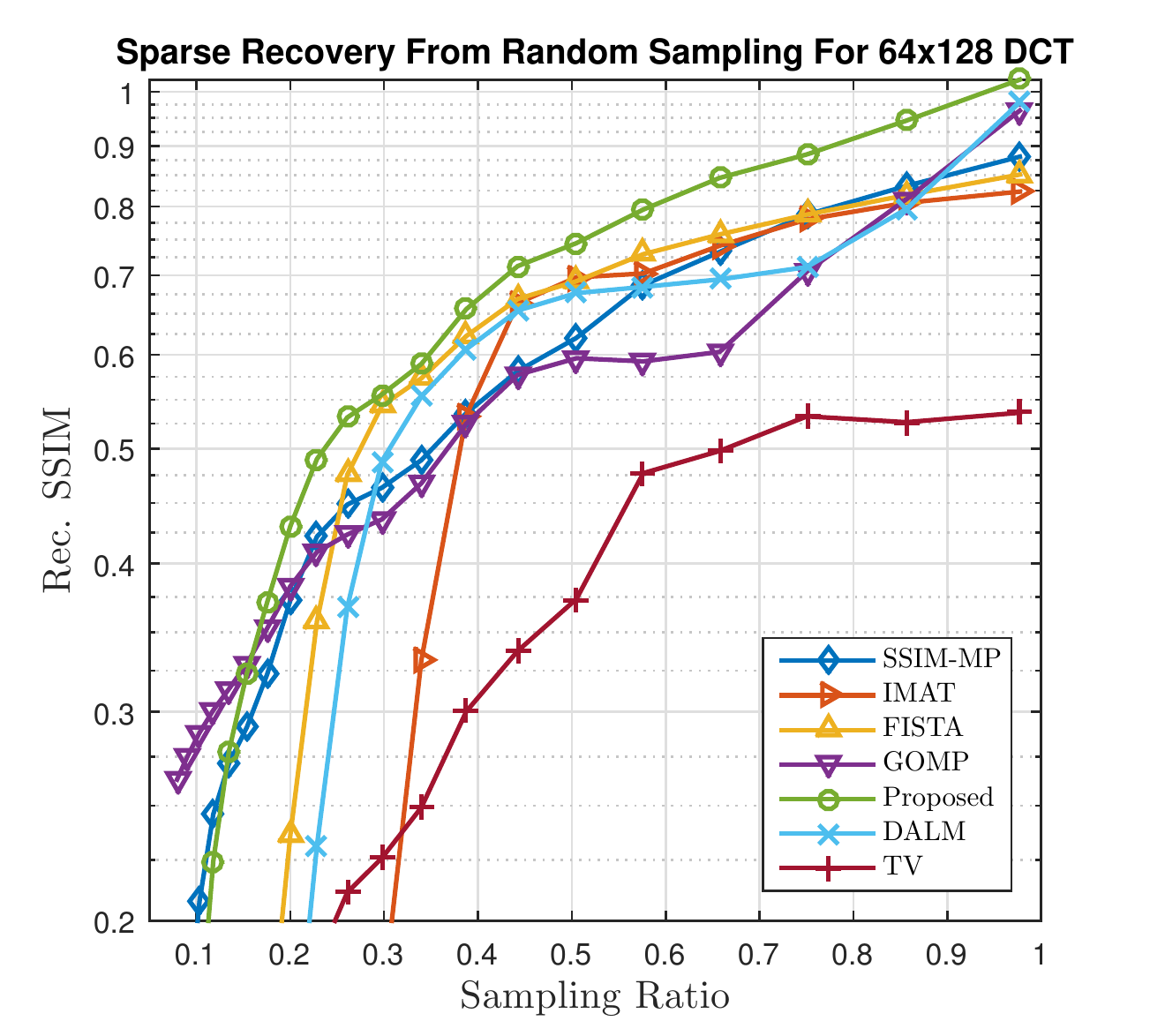}%
\label{fig_1:b}} \hspace{-0.4cm}
\subfloat[PSNR (DWT 128)]{\includegraphics[width=2.0in]{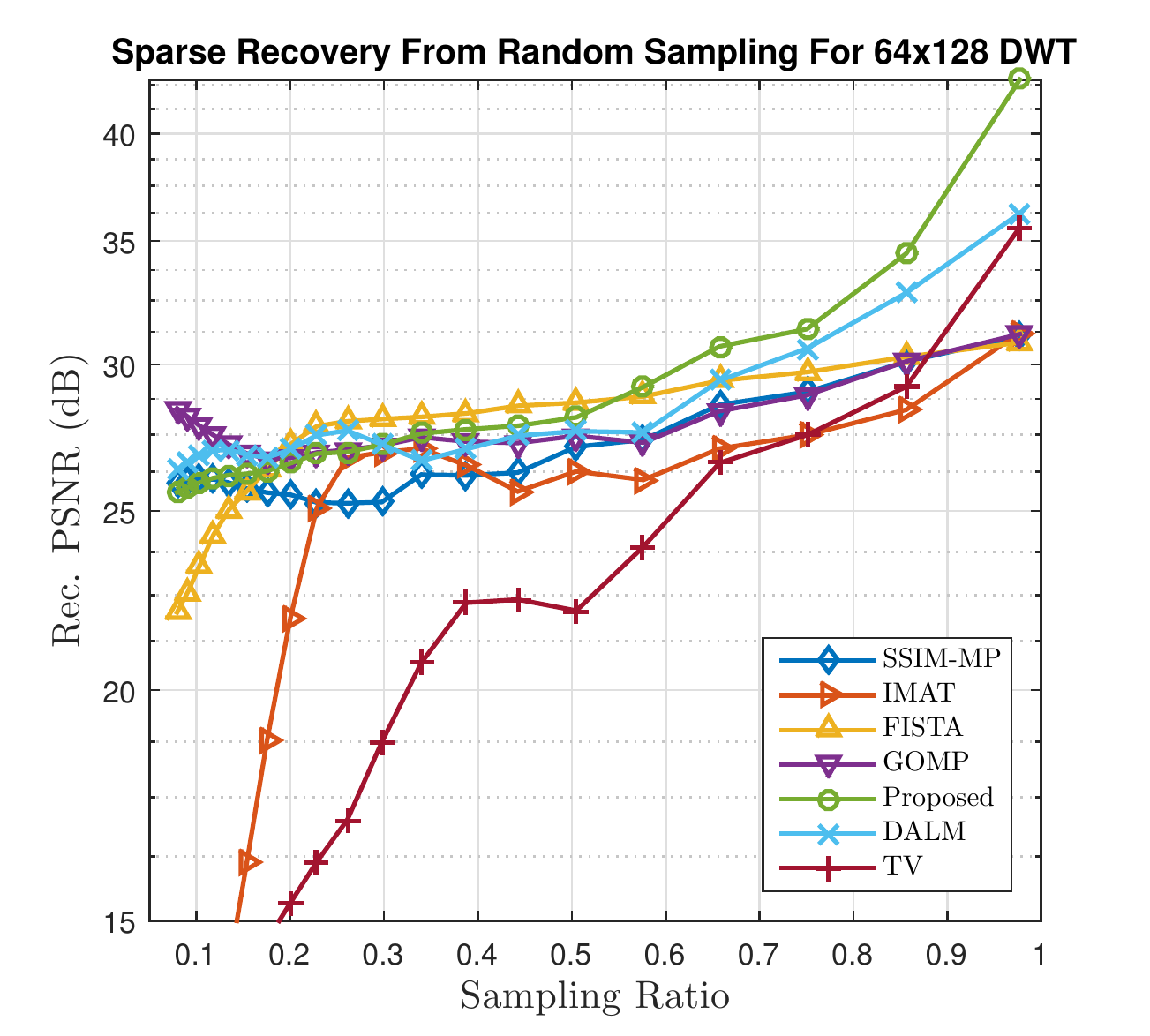}%
\label{fig_1:c}\hspace{-0.3cm}}  
\subfloat[SSIM (DWT 128)]{\includegraphics[width=2.0in]{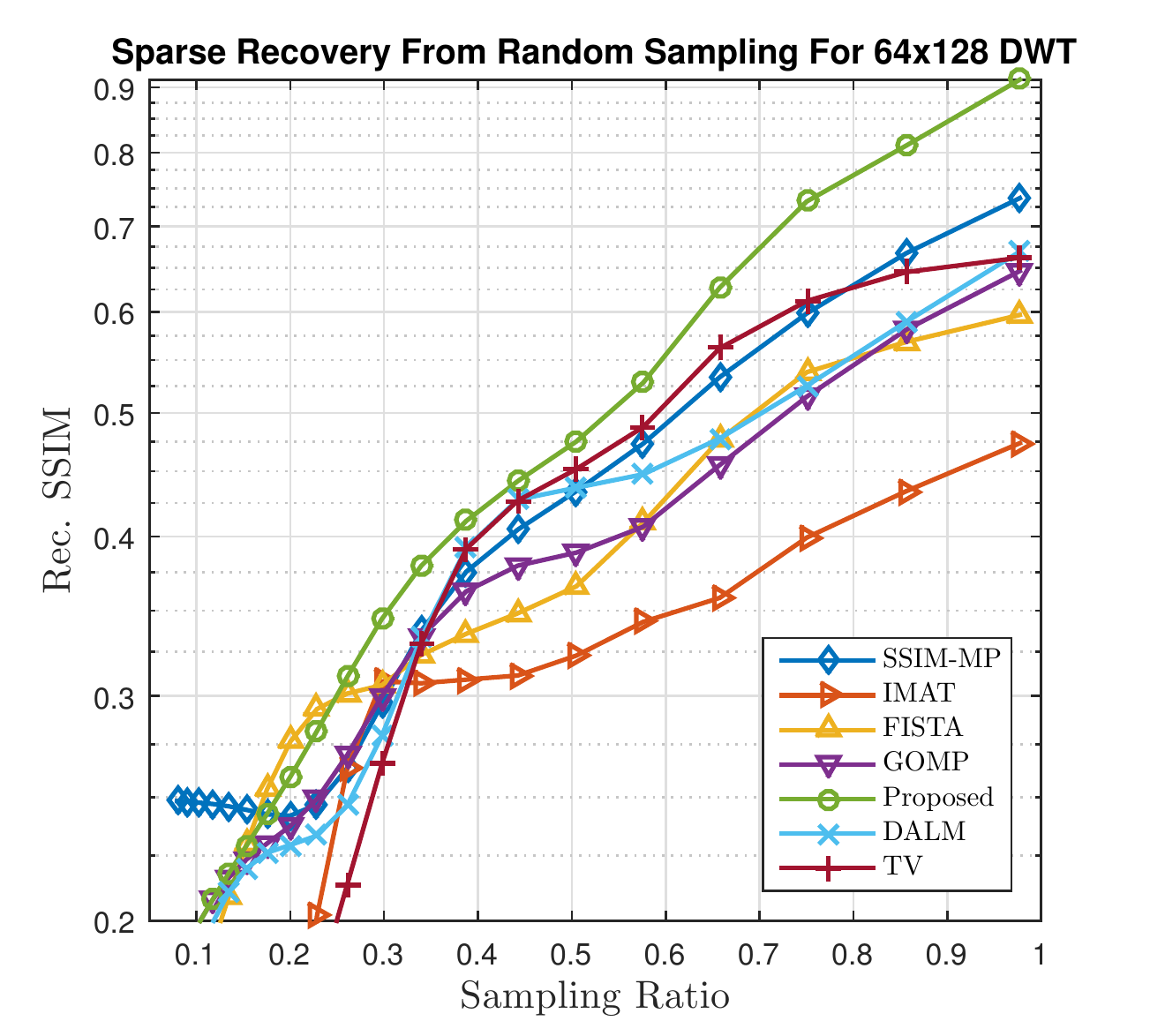}%
\label{fig_1:d}}
\caption{Quality performance of sparse recovery methods versus the rate of sampling of $64\times 1$ image vectors. In  (\ref{fig_1:a}) and (\ref{fig_1:b}) we have used $64\times 128$ DCT and in (\ref{fig_1:c}) to (\ref{fig_1:d}) we have incorporated $64\times 128$ DWT atoms for sparse approximation.}
\label{fig_1}
\end{figure*}

\begin{table*}[!t]
\vspace{-0.2cm}
\caption{Performance comparison of several iterative methods for missing sample recovery of 2D image. The pairs are (PSNR($\mathrm{dB}$), SSIM)}
\vspace{-0.2cm}
\begin{tabular}{|l|l||p{1.85cm}|p{1.85cm}|p{1.85cm}|p{1.85cm}|p{1.85cm}|p{1.85cm}|p{1.85cm}|} \hline 
& & & & & & & & \\[-1.5ex]
 & $sr$ & IMATI (DCT) & Hosseini et, al. (DCT) & Guleryuz (DCT 8)& EM 
(Curvelet) & MCA (Curvelet) & Proposed (DCT) & Proposed (Curvelet) \\ \hline \hline 
 \multirow{3}{*}{Barbara} & $0.1$ & 15.4129~ 0.2871 & 20.6111~ 0.5255 & 15.6349~ 0.3037 & \textcolor{red}{ \bf 20.7160}~ 0.4939 & 20.4960~ 0.5163 & 20.1001~ 0.4320 & 19.8686~ \textcolor{red}{ \bf 0.5313} \\ \cline{2-9}
 & $0.3$ & 19.2285~ 0.5667 & 22.8145~ 0.6468 & 16.3647~ 0.3602 & 24.1875~ 0.7071 & \textcolor{red}{ \bf 25.2570}~ 
\textcolor{red}{ \bf 0.7976} & 22.7122~ 0.6156 & 24.3881~ 0.7872 \\ \cline{2-9}
 & $0.5$ & 20.9206~ 0.6833 & 24.8344~ 0.7566 & 27.3870~ 0.8823 & 26.7496~ 
0.8207 & 28.2209~ 0.8948 & 25.0435~ 0.7551 & \textcolor{red}{ \bf 28.2862}~ \textcolor{red}{ \bf 0.8966} \\ \hline \hline 
 \multirow{3}{*}{Lena}  & $0.1$ & 16.9363~ 0.3573 & 22.9252~ 0.6420 & 15.2763~ 0.3385 & 21.8212~ 
0.6232 & 21.7169~ 0.6359 & 23.1364~ 0.6590 & \textcolor{red}{ \bf 23.5069}~ \textcolor{red}{ \bf 0.6996} \\ \cline{2-9}
 & $0.3$ & 21.0841~ 0.6204 & 26.2759~ 0.7608 & 15.9784~ 0.3480 & 26.1313~ 
0.7859 & 27.9148~ 0.8638 & 26.5411~ 0.7624 & \textcolor{red}{ \bf 28.2068}~ \textcolor{red}{ \bf 0.8687} \\ \cline{2-9}
 & $0.5$ & 23.2964~ 0.7575 & 28.4910~ 0.8313 & 29.0925~ 0.8960 & 28.8726~ 
0.8606 & 31.4543~ 0.9274 & 30.0204~ 0.8735 & \textcolor{red}{ \bf 31.6610}~ \textcolor{red}{ \bf 0.9286} \\ \hline \hline 
  \multirow{3}{*}{House}  & $0.1$ & 13.0318~ 0.2269 & 23.9295~ 0.5760 & 13.0612~ 0.2340 & 23.5729~ 
0.6940 & 22.9949~ 0.7041 & 23.9442~ 0.5870 & \textcolor{red}{ \bf 25.9459}~ \textcolor{red}{ \bf 0.7122} \\ \cline{2-9}
 & $0.3$ & 17.3905~ 0.4491 & 28.6938~ 0.7869 & 13.7061~ 0.1838 & 28.6960~ 
0.7940 & 30.8723~ \textcolor{red}{ \bf 0.8689} & 28.9419~ 0.7886 & \textcolor{red}{ \bf 31.1493}~ 0.8539 \\ \cline{2-9}
 & $0.5$ & 19.0919~ 0.5689 & 31.0684~ 0.8594 & 28.9862~ 0.8775 & 31.1644~ 
0.8357 & 34.0413~ \textcolor{red}{ \bf 0.9176} & 32.1972~ 0.8819 & \textcolor{red}{ \bf 34.2095}~ 0.9140 \\ \hline \hline 
 \multirow{3}{*}{Peppers}  & $0.1$ & 15.2167~ 0.3175 & 22.2164~ 0.6636 & 13.8936~ 0.2744 & 21.2991~ 
0.6333 & 20.5362~ 0.6189 & 22.3154~ 0.6600 & \textcolor{red}{ \bf 22.3869}~ \textcolor{red}{ \bf 0.6858} \\ \cline{2-9}
 & $0.3$ & 19.7079~ 0.6175 & 25.3842~ 0.7645 & 14.6123~ 0.2664 & 25.8395~ 
0.7911 & 27.0907~ 0.8512  & 25.5390~ 0.7680 & \textcolor{red}{ \bf 27.3406}~  \textcolor{red}{ \bf 0.8535}  \\ \cline{2-9} 
 & $0.5$ & 21.6589~ 0.7299 & 27.6993~ 0.8350 & 28.0811~ 0.8912 & 28.3704~ 
0.8561 & 30.3404~ 0.9130 & 28.5436~ 0.8591 & \textcolor{red}{ \bf 30.5844}~ \textcolor{red}{ \bf 0.9137} \\ \hline 
\end{tabular} 
\label{Table_1}
\end{table*}

\begin{table*}[!h]
\vspace{-0.2cm}
\centering
\caption{Running time(s) of several iterative methods for missing sample recovery of 2D image }
\vspace{-0.2cm}
\begin{tabular}{|l||p{1.7cm}|p{1.7cm}|p{1.7cm}|p{1.7cm}|p{1.7cm}|p{1.7cm}|p{1.7cm}|} \hline
& & & & & & & \\[-1.5ex]
 $sr$ & IMATI (DCT) & Hosseini et, al. (DCT) & Guleryuz (DCT 8) & EM 
(Curvelet) & MCA (Curvelet) & Proposed (DCT) & Proposed (Curvelet) \\ \hline \hline 
  $0.1$ & 19.6828 & 0.8137 & 187.9410 & 89.5931 & 95.3843 & 1.4531 & 64
.6750 \\ \hline 
  $0.3$ & 15.7252 & 0.8621 & 87.8616 & 42.1668 & 98.3200 & 1.2188 & 64
.6035 \\ \hline 
  $0.5$ & 10.8503 & 0.9567 & 74.4042 & 27.9614 & 94.3267 & 1.1563 & 64
.9105 \\ \hline 
\end{tabular}
\label{Table_2}
\end{table*}

\begin{figure*}[!t]
\vspace{-0.3cm}
\centering
\subfloat[]{\includegraphics[width=1.25in]{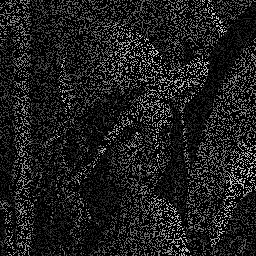}%
\label{fig_2:a}}  
\subfloat[]{\includegraphics[width=1.25in]{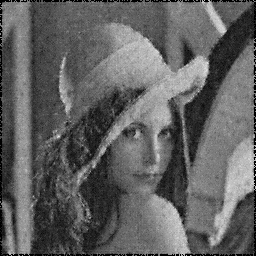}%
\label{fig_2:b}} 
\subfloat[]{\includegraphics[width=1.25in]{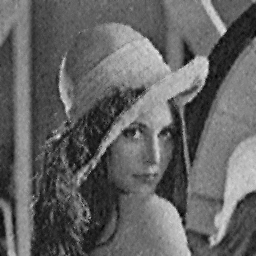}%
\label{fig_2:c}}  
\subfloat[]{\includegraphics[width=1.25in]{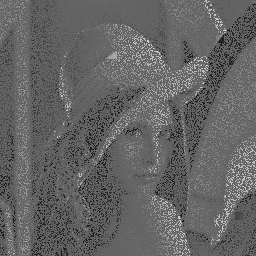}%
\label{fig_2:d}} \\[-2ex]
\subfloat[]{\includegraphics[width=1.25in]{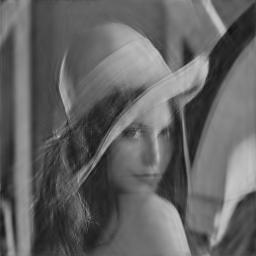}%
\label{fig_2:e}}  
\subfloat[]{\includegraphics[width=1.25in]{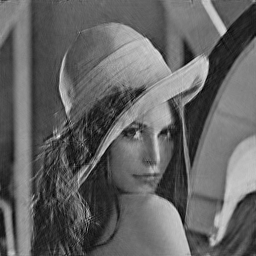}%
\label{fig_2:f}} 
\subfloat[]{\includegraphics[width=1.25in]{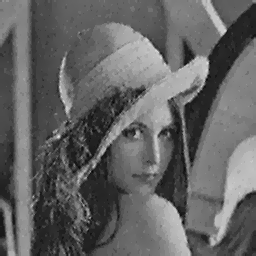}%
\label{fig_2:g}}  
\subfloat[]{\includegraphics[width=1.25in]{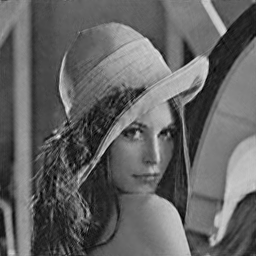}%
\label{fig_2:h}}
\caption{Visual quality of the reconstructed image Lena from 0.3 random samples, From top to bottom and left to right: (\ref{fig_2:a}) Image with missing samples, (\ref{fig_2:b}) Rec. Image via IMATI, (\ref{fig_2:c}) Rec. Image via Hosseini et, al., (\ref{fig_2:d}) Rec. Image via Guleryuz, (\ref{fig_2:e}) Rec. Image via EM (Curv.), (\ref{fig_2:f}) Rec. Image via MCA (Curv.), (\ref{fig_2:g}) Rec. Image via Proposed (DCT), (\ref{fig_2:h}) Rec. Image via Proposed (Curv.). For PSNR and SSIM values refer to Table \ref{Table_1}.} 
\label{fig_2}
\end{figure*}

\section{Simulation Results}
\label{sec:Simulations}
\subsection{Sparse vector recovery}
\label{subsec:Exp1}

In this experiment, we compare the quality performance of the proposed method for recovery of missing samples of image patches with some popular sparse recovery algorithms. We use IMAT\footnote{{http://ee.sharif.edu/$\sim$imat/}}, DALM\footnote{{{https://people.eecs.berkeley.edu/$\sim$yang/software/l1benchmark/}}} \cite{Yang11}, TV\footnote{{http://www.caam.rice.edu/$\sim$optimization/L1/TVAL3/}} \cite{TV013}, FISTA \cite{Beck09},  GOMP\footnote{{http://islab.snu.ac.kr/paper/gOMP.zip}} \cite{GOMPWang} and the method in \cite{Oga13} which we call it SSIM-based Matching Pursuit  (SSIM-MP). For simulations of this part, we extract $8\times8$ patches of sample gray-scale images. We then vectorize the patches using raster scanning and select 50 patch vectors at random. From each patch, we take $m$ samples, chosen uniformly at random, and the sampling ratio of the signal defined as $sr=\frac{m}{n}$ varies between $(0,1)$. We use over-complete ($64\times 128$) DCT and DWT\footnote{MATLAB command \texttt{wmpdictionary(64,'lstcpt',\{'wpsym4',4\})}} dictionaries for reconstruction. Since the exact sparsity is unknown, to use matching pursuit methods we assume the signal is $10\%$ sparse. After the sparse recovery of missed samples, we then average over random experiments and plot the PSNR and SSIM versus the sampling rate. The parameters for TV and FISTA are set to their default (source code) values. The values of the exponential threshold parameters in IMAT are set to $\alpha=0.2$, $\beta = 0.2 \norm \Db^T \Hb^T\yvec \norm_\infty$ and $\lambda=0.5$. The stopping criterion for DALM and L1-LS are set to minimum duality gap. We choose maximum iteration count of 50 as the stopping criterion for the remaining algorithms. The parameters for our proposed method are chosen as $\sigma_1 =2 \frac{m}{n}= 2 sr $, $\mu=0.8$, $\zeta=0.2$, $K_0=n-1=63$ and $\rho = 1.1$. Also similar to FISTA, The value of $\alpha_{\min}$ is  set to $10 ^{-4}$. As depicted in Fig.~\ref{fig_1}, the proposed algorithm mostly outperforms the competing algorithms and gives a better reconstruction quality compared to DALM and TV which commonly solve the \lone-optimization problem using ADMM. The SSIM performance as given in Fig.~\ref{fig_1} also confirms this efficiency. 
\subsection{Image Completion}
\label{subsec:Exp2}
In this part, we compare the performance of the proposed algorithm for 2D image reconstruction with several iterative methods, namely IMATI \cite{Azgh13}, Hosseini et, al. \cite{Hosseini14} and the well-know inpainting algorithms including Guleryuz \cite{Guler06}, MCA \cite{Elad05} and EM \cite{Fadili07}. The parameters for IMATI are set the default values $\lambda=1.8$, $\mathrm{iter}_{\max}=100$ and $\epsilon=1e-4$. The method of Hosseini et, al. is simulated via DCT lowpass filtering method with 10 iterations. The Guleryuz method has been run for 100 iterations using DCT 16 transform. The parameters of EM are set to $\lambda=10$, $\sigma=1$ and $\epsilon=1e-3$ and for MCA we choose $\mathrm{iter}_{\max}=100$ and $\lambda_\text{stop}=1e-4$.  The parameters of our proposed 2D algorithm are also chosen similar to the 1D case except that we use $K_0=2.5(N-1)$, $\sigma=6 sr$, $\lambda=1.2$ and $\mathrm{iter}_{\max}=40$. 
 Table \ref{Table_1} shows the reconstruction PSNR and SSIM values for some test images at sampling rates 0.1, 0.3 and 0.5. We have used both DCT and Curvelet transforms for reconstruction with our algorithm. The values in Table \ref{Table_1} which are (PSNR, SSIM) pairs, confirm the quality performance of the proposed method compared to other inpainting algorithms in most cases. Furthermore, comparing the running time of the proposed algorithm (with DCT transform) with other methods, as given in Table \ref{Table_2}, implies its relatively low computational complexity.

\section{Conclusion}
\label{sec:conclusion}
In this paper, we proposed an iterative method for missing sample recovery of image signals using sparse approximation. In particular, we proposed an algorithm based on \lone-minimization for missing sample recovery of 1D image patch vectors. We incorporated the Convex SIMilarity (CSIM) index, which similar to MSE, is well suited for mathematical manipulations and like SSIM, benefits some sense of error-visibility feature. The optimization problem incorporating this fidelity metric is then solved via ADMM. We also introduced a 2D variant of the proposed method which can directly be used to inpaint the whole corrupted image without need to extract and vectorize small patches. Simulation results approve the performance quality of the proposed algorithm for 1D and 2D image completion.


%
%



%

\end{document}